\documentclass[11pt]{article}

\usepackage{acl}

\usepackage{times}
\usepackage{latexsym}

\usepackage[normalem]{ulem}
\usepackage{color}
\usepackage{xcolor}

\usepackage[T1]{fontenc}

\usepackage[utf8]{inputenc}

\usepackage{microtype}

\usepackage{inconsolata}

\usepackage{graphicx}
\usepackage{booktabs}
\usepackage{multirow}
\usepackage[table]{xcolor}
\usepackage{makecell}
\usepackage{amsmath}
\usepackage{amssymb}
\usepackage{algorithm}
\usepackage{algpseudocode}
\usepackage{balance}
\usepackage{subcaption}
\usepackage{caption}
\title{C-MORAL: Controllable Multi-Objective Molecular Optimization with Reinforcement Alignment for LLMs}

\author{
  Rui Gao$^{1\ast}$ \quad Youngseung Jeon$^{1}$\thanks{These authors contributed equally to this work.} \quad Swastik Roy$^{2}$ \quad Morteza Ziyadi$^{2\dagger}$ \quad Xiang `Anthony' Chen$^1$\thanks{These authors jointly supervised this work.} \\
  $^{1}$University of California, Los Angeles \quad $^{2}$Amazon \\
  \texttt{\{rgao727, ysj, xac\}@ucla.edu} \\
  \texttt{\{roswasti, mziyadi\}@amazon.com}
}

\begin{document}

\maketitle
\begin{abstract}
Large language models (LLMs) show promise for molecular optimization, but aligning them with selective and competing drug-design constraints remains challenging. We propose \textsc{C-Moral}, a reinforcement learning post-training framework for controllable multi-objective molecular optimization. \textsc{C-Moral} combines group-based relative optimization, property score alignment for heterogeneous objectives, and continuous non-linear reward aggregation to improve stability across competing properties. Experiments on the C-MuMOInstruct benchmark show that \textsc{C-Moral} consistently outperforms state-of-the-art models across both in-domain and out-of-domain settings, achieving the best Success Optimized Rate (SOR) of 48.9\% on IND tasks and 39.5\% on OOD tasks, while largely preserving scaffold similarity. These results suggest that RL post-training is an effective way to align molecular language models with continuous molecular design objectives. Our code and models are publicly available at \url{https://github.com/Rwigie/C-MORAL}.
\end{abstract}

\begin{figure*}[t]
  \centering
  \includegraphics[width=\textwidth, height=0.9\textheight, keepaspectratio]{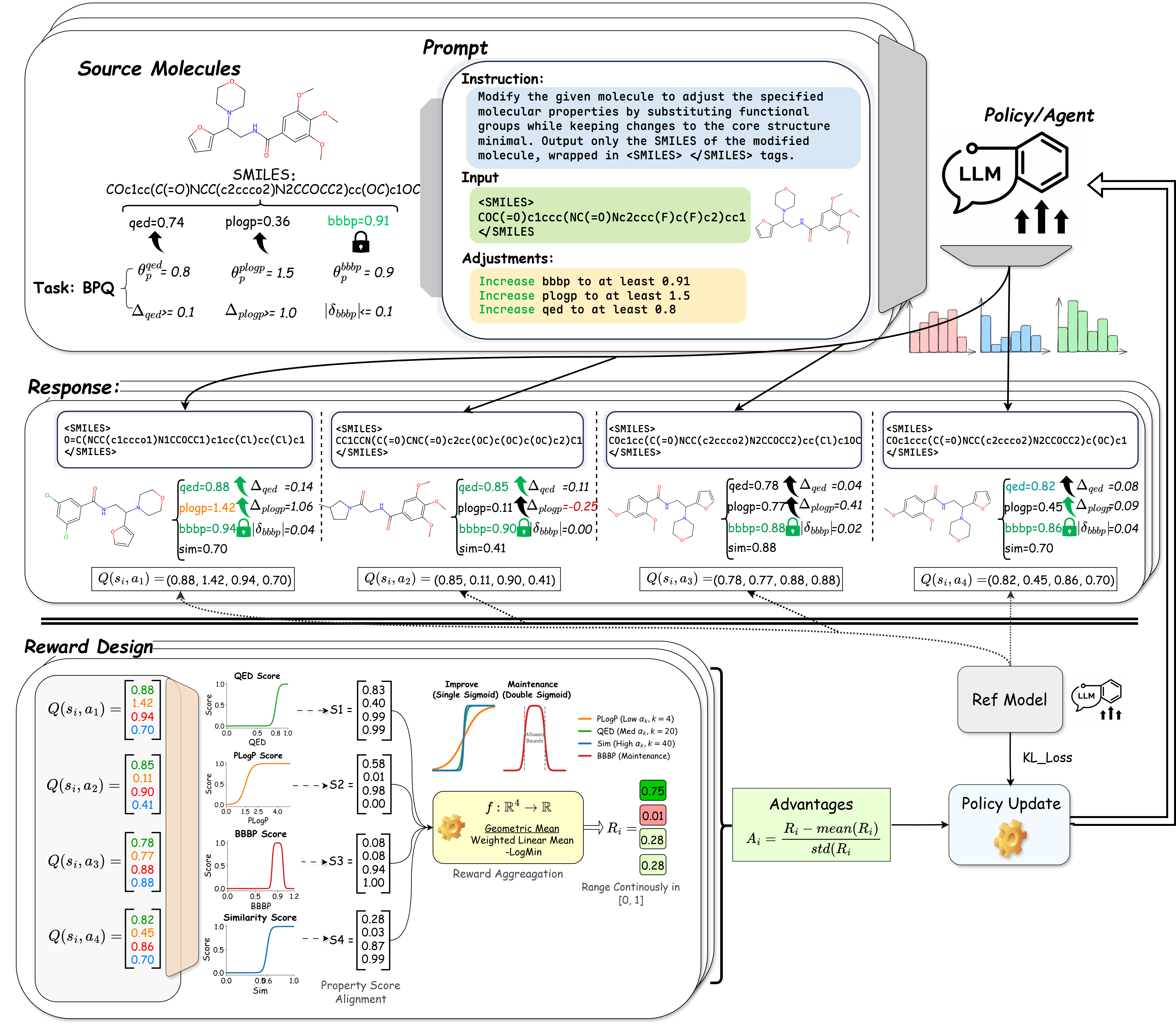}
  \caption{Overview of \textsc{C-MORAL} generation, training pipeline}
  \label{fig:framework}
\end{figure*}

\section{Introduction}

Drug discovery is a lengthy and costly process, with the hit-to-lead transition and the lead optimization stage being particularly critical \cite{katsuno2015hit, sadybekov2023computational}. This process is inherently a multi-objective optimization (MOO) problem \cite{guan2019admet}. 

Practical molecular optimization in drug discovery is rarely a matter of simply improving all properties \cite{fialkova2021libinvent}. In many cases, a lead compound already approaches desirable levels in a subset of attributes; therefore, optimization requires selective improvement of the remaining deficient properties while maintaining those that already meet the target criteria, and simultaneously preserving the core scaffold and synthetic feasibility \cite{zhou2024decompoptcontrollabledecomposeddiffusion}. For example, ControllableGPT \cite{liu2025controllablegpt} aims to increase binding affinity while preserving the core scaffold by making minimal, targeted edits that remove liabilities. Such tasks have been formalized as a benchmark setting for controllable multi-property, multi-objective optimization (C-MuMO) \cite{dey2025large}. C-MuMO highlights the need to specify property-wise objectives and thresholds for either improvement or preservation, thereby better reflecting the fine-grained requirements of real medicinal chemistry. In this context, ``controllable'' means that the model must follow explicit, property-wise directives---improving only the deficient attributes while preserving those already satisfying target criteria---instead of indiscriminately maximizing a single aggregate objective.



Despite its practical importance, controllable molecular optimization remains challenging. The difficulty lies not in improving molecular properties alone, but in doing so under explicit, property-wise requirements: a model must improve deficient attributes while preserving those already meeting target criteria, without violating scaffold or feasibility constraints. Existing methods struggle to satisfy such fine-grained control reliably.
RL-based approaches typically optimize a scalarized reward over multiple objectives \cite{wang2024multi, park2025mol}, which can lead to brittle trade-offs under strict and competing constraints. LLM-based approaches adapted by SFT or prompting can leverage strong chemical priors \cite{ye2023drugassistlargelanguagemodel}, but they often treat chemical constraints as soft instructions rather than hard requirements. As a result, they may achieve apparent score improvements through undesired edits, such as altering the core scaffold or violating preservation constraints \cite{guo2023largelanguagemodelschemistry}.

To achieve genuine, controllable optimization and overcome the limitations of both traditional RL and SFT paradigms, we introduce \textsc{C-Moral}, a novel Reinforcement Learning post-training framework designed specifically for molecular LLMs. An overview of the framework is shown in Figure \ref{fig:framework}. \textsc{C-Moral} bridges the gap between the robust chemical priors of LLMs and the strict, multi-objective demands of lead optimization. The framework aligns the LLM with rigorous scientific constraints through group-aware relative feedback. Specifically, we adapt two distinct algorithms: Group Relative Policy Optimization (GRPO) \cite{shao2024deepseekmathpushinglimitsmathematical} and Group reward-Decoupled Normalization Policy Optimization (GDPO) \cite{liu2026gdpogrouprewarddecouplednormalization} to create two variants of our framework that evaluate candidate molecules against their peers generated from the same prompt. By bypassing the need for a separate, memory-intensive value network, these group-relative approaches significantly enhance training efficiency and memory utilization.

Within the framework, we propose two distinct algorithmic variants to handle competing properties: \textsc{C-Moral}-GRPO, which utilizes a sigmoid-aligned geometric mean reward to inherently penalize property bottlenecks, and \textsc{C-Moral}-GDPO, which decouples conflicting objectives using a smooth minimum (-Log-Sum-Exp) advantage to prevent the implicit sacrifice of any attribute.

Evaluated on the stringent C-MuMOInstruct benchmark \cite{dey2025large}, our framework significantly outperforms state-of-the-art baselines. In summary, our main contributions are:
\begin{itemize}
    \setlength{\itemsep}{0pt}
    \setlength{\parskip}{0pt}
    \setlength{\parsep}{0pt}
    \item We propose \textsc{C-Moral}, a reinforcement learning post-training framework that enables precise multi-objective alignment for molecular LLMs.
    \item To ensure stable and controllable optimization, we introduce a group-relative RL strategy (incorporating \textsc{GRPO} and \textsc{GDPO}) coupled with a novel reward design that unifies metric scales and enforces joint constraint satisfaction.
    \item \textsc{C-Moral} achieves a best success optimized rate of 48.9\% on in-domain tasks and 39.5\% on out-of-domain tasks on the C-MuMO benchmark, consistently outperforming strong instruction-tuned baselines while preserving scaffold similarity.
\end{itemize}

\section{Related Work}\label{sec:related-work}

\subsection{Group-based Policy Optimization}
Compared with PPO, which relies on an additional critic network, GRPO removes the value model by normalizing rewards within a sampled candidate group, thereby reducing computational overhead and simplifying optimization \cite{shao2024deepseekmathpushinglimitsmathematical}. More recently, GDPO further extends this idea by decoupling advantage normalization across reward dimensions, which is especially suitable for settings with strongly conflicting objectives \cite{liu2026gdpogrouprewarddecouplednormalization}. Our work builds on this line of research, but adapts it to molecular optimization by combining group-relative policy updates with property score alignment and non-linear reward aggregation under heterogeneous molecular constraints.

\subsection{Molecular Optimization}
Computational approaches for molecular optimization have evolved from heuristic search to reinforcement learning (RL) and deep generative modeling \cite{gomez2018automatic,zhou2019optimization,walters2020applications,olivecrona2017molecular,wang2023retrievalbasedcontrollablemoleculegeneration}. Existing methods explore diverse action spaces, including token-level SMILES generation \cite{weininger1988smiles}, graph construction \cite{erikawa2023gargoyles}, and structure refinement in continuous coordinates \cite{barrett2024reinforcement}. Despite their effectiveness, these approaches often struggle in complex multi-objective settings due to inefficient exploration and brittle reward design, especially when competing objectives and hard constraints must be satisfied simultaneously \cite{brown2019guacamol,wang2024multi}.

Large language models (LLMs) have recently emerged as a promising alternative for molecular generation and optimization by treating chemical representations such as SMILES as a language \cite{bagal2021molgpt,ghugare2023searchinghighvaluemoleculesusing}. Recent work has extended LLMs toward broader molecular understanding and optimization, including molecular assistants such as LLaMo \cite{park2024llamolargelanguagemodelbased}, conversational drug editing frameworks such as ChatDrug \cite{liu2023chatgptpoweredconversationaldrugediting}, and optimization-oriented methods such as LICO and MOLLEO \cite{nguyen2025licolargelanguagemodels,wang2025efficientevolutionarysearchchemical}. REPO further explores GRPO-style optimization for improving reasoning in instruction-based molecular generation \cite{li2026reference}. However, existing approaches still provide limited study of controllable multi-objective molecular optimization, particularly when multiple continuous and conflicting objectives must be satisfied while preserving structural fidelity. Our work instead focuses on aligning molecular LLMs with such settings through reward-aware RL post-training.

\section{Method}

\subsection{Problem Formulation}

We formulate Controllable Multi-objective Molecule Optimization (C-MuMO) as a constrained generation task. Given an initial sub-optimal ``hit'' molecule $M_x$ and a natural language instruction $I$, the objective is to generate an optimized ``lead'' molecule $M_y$ that selectively improves specific molecular properties up to pharmaceutically relevant levels, while maintaining others that already meet the criteria \cite{dey2025large}.

Let $\mathcal{P}$ denote the set of target pharmacological properties. For each property $p \in \mathcal{P}$, we define a pharmaceutically relevant threshold $\Theta_p$ and an improvement/stability margin $\Delta_p$. Based on the initial molecule $M_x$, the properties are partitioned into two disjoint subsets: 
\begin{itemize}
    \item Sub-optimal properties ($\mathcal{P}_i$): Properties that require targeted enhancement, defined as $\mathcal{P}_i = \{p \in \mathcal{P} \mid p(M_x) \mbox{ is worse than } \Theta_p\}$.
    \item Near-optimal properties ($\mathcal{P}_s$): Properties that already meet the criteria and must be preserved, defined as $\mathcal{P}_s = \{p \in \mathcal{P} \mid p(M_x) \mbox{ is better than or equal to } \Theta_p\}$.
\end{itemize}

Detailed values of $\Delta_p$ and $\Theta_p$ for different properties are given in Table \ref{tab:dataset_summary}.

The generation policy $\pi_\Theta(M_y \mid M_x, I)$, parameterized by a Large Language Model (LLM), is tasked with generating $M_y$ such that it strictly satisfies three non-differentiable constraints simultaneously: (1) similarity constraint, ensuring the structural similarity between $M_x$ and $M_y$ (e.g., Tanimoto similarity \cite{bajusz2015tanimoto}) remains above a predefined threshold; (2) improvement constraint, where every sub-optimal property $p \in \mathcal{P}_i$ must exhibit an absolute improvement of at least $\Delta_p$ in the desired direction (i.e., $|p(M_y) - p(M_x)| \ge \Delta_p$); and (3) stability constraint, which strictly bounds the absolute deviation of every near-optimal property $p \in \mathcal{P}_s$ to prevent catastrophic degradation (i.e., $|p(M_y) - p(M_x)| \le \Delta_p$).

\subsection{Property Score Sigmoid Alignment}

Molecular properties inherently exist on vastly different scales. Directly aggregating raw values creates severe numerical instability. To address this, we introduce a property-specific score shaping mechanism that normalizes all metrics into a unified $[0, 1]$ preference space. 

Let $\sigma(x) = 1 / (1 + \exp(-x))$ denote the standard sigmoid function. For brevity, let $v_p$ denote the realized property value $v_p(M_y)$ of the generated molecule. For sub-optimal properties ($\mathcal{P}_i$), we set the target threshold $T_p$ (e.g., the baseline plus $\Delta_p$) as the midpoint. The improvement score $s^{imp}_p$ is:
\begin{equation}
  \label{eq:sigmoid_imp}
  s^{imp}_p = \sigma(\alpha_p \cdot (v_p - T_p))
\end{equation}
where $\alpha_p$ controls the steepness. Crucially, we parameterize $\alpha_p$ to be inversely proportional to the property-specific margin $\Delta_p$ (i.e., $\alpha_p \propto 5/\Delta_p$). This adaptive scaling ensures that achieving the required $\Delta_p$ for any metric---regardless of its original numerical scale---maps to an equivalent reward magnitude (e.g., exactly 0.5 at $T_p$). It effectively prevents metrics with naturally broader scales from dominating the gradient updates.

Conversely, for near-optimal properties ($\mathcal{P}_s$), the property must be constrained within a strict tolerance band $[L_p, U_p]$, where $L_p = p(M_x) - \Delta_p$ and $U_p = p(M_x) + \Delta_p$. To heavily penalize deviations, we design a Double Sigmoid score function:
\begin{equation}
  \label{eq:double_sigmoid}
  s^{stab}_p = \sigma(\alpha_p \cdot (U_p - v_p)) \cdot \sigma(\alpha_p \cdot (v_p - L_p))
\end{equation}
This effective formulation creates a high-reward ``plateau'' within the acceptable range and imposes exponential decay upon any boundary violation.
\subsection{Reward Aggregation}
 After getting the score, we aggregate these shaped scores using the geometric mean to enforce the concurrent satisfaction of all constraints. Let $N = |\mathcal{P}_i| + |\mathcal{P}_s|$. The total reward $R_{total}(M_y)$ is:
\begin{equation}
  \label{eq:geometric_mean}
  R_{total}(M_y) = \left( \prod_{p \in \mathcal{P}_i} s^{imp}_p \prod_{q \in \mathcal{P}_s} s^{stab}_q \right)^{\frac{1}{N}}
\end{equation}
Here, we deliberately employ the geometric mean rather than the standard arithmetic mean (linear scalarization) because it can be viewed as a continuous approximation of a $min$ operator. The full explanation is in Appendix \ref{appendix:aggregation}.

\subsection{GRPO Optimization}
For a given prompt, the policy $\pi_\Theta$ samples a group of $G$ candidates $\mathcal{G} = \{y_1, \dots, y_G\}$. For each $y_i$, we compute its holistic reward $R_i = R_{total}(y_i)$. We then normalize these rewards to compute the relative advantage $A_i$:
\begin{equation}
  \label{eq:grpo_advantage}
  A_i^{GRPO} = \frac{R_i - \mu_{\mathcal{G}}}{\sigma_{\mathcal{G}} + \epsilon}
\end{equation}
where $\mu_{\mathcal{G}}$ and $\sigma_{\mathcal{G}}$ are the mean and standard deviation of rewards within $\mathcal{G}$.

To optimize the policy, we first define the clipped surrogate objective $J_i(\Theta)$ for each candidate, where $\rho_i = \pi_\Theta(y_i) / \pi_{ref}(y_i)$ is the importance ratio:
\begin{equation}
  \label{eq:grpo_surrogate}
  J_i(\Theta) = \min \left( \rho_i A_i^{GRPO}, \mbox{clip}(\rho_i, 1-\epsilon_c, 1+\epsilon_c) A_i^{GRPO} \right)
\end{equation}

The final \textsc{C-Moral}-GRPO policy follows the same loss $\mathcal{L}_{GRPO}(\Theta)$ employing an explicit KL divergence penalty to prevent catastrophic deviation from the reference model $\pi_{ref}$:
\begin{equation}
  \label{eq:grpo_loss}
  \mathcal{L}_{GRPO}(\Theta) = - \frac{1}{G} \sum_{i=1}^{G} \left[ J_i(\Theta) - \beta \mathbb{D}_{KL}(\pi_\Theta \| \pi_{ref}) \right]
\end{equation}
By decoupling the absolute reward magnitude, \textsc{C-Moral}-GRPO inherently preserves structural diversity and circumvents policy peaking. Detailed implementation is provided in Appendix \ref{appendix:grpo_implementation}.

\subsection{GDPO Optimization}
To explicitly disentangle conflicting multi-objective feedback, we propose \textsc{C-Moral}-GDPO, which evaluates relative superiority independently for each property before aggregating them into a unified preference signal.

For a generated group $\mathcal{G} = \{y_1, \dots, y_G\}$, we first compute a property-specific relative advantage for each candidate $y_i$ and each property $p \in \mathcal{P}_{\text{total}}$. Let $r_{p,i}$ denote the shaped reward of candidate $y_i$ on property $p$. The decoupled advantage is defined as
\begin{equation}
  \label{eq:gdpo_decoupled_adv}
  A_{p,i} = \frac{r_{p,i} - \mu_{p,\mathcal{G}}}{\sigma_{p,\mathcal{G}} + \epsilon},
\end{equation}
where $\mu_{p,\mathcal{G}}$ and $\sigma_{p,\mathcal{G}}$ denote the group mean and standard deviation for property $p$.

To aggregate these decoupled advantages, we use a Log-Sum-Exp form as a smooth approximation that emphasizes the lowest-performing objective:
\begin{equation}
  \label{eq:gdpo_softmin}
  A_i^{\mathrm{GDPO}} = - \log \left( \sum_{p \in \mathcal{P}_{\text{total}}} \exp(-A_{p,i}) \right).
\end{equation}
The resulting aggregated advantage is then used in the policy optimization objective in Equation \ref{eq:grpo_loss}. In this way, \textsc{C-Moral}-GDPO reduces the risk that easily optimized properties dominate training while stricter constraints are implicitly sacrificed.
Detailed implementation of GDPO is in Appendix \ref{appendix:gdpo_implementation}.

Further analysis of the reward aggregation choices is provided in Appendix \ref{appendix:gm_vs_logmin}.

\section{Experimental Setup}
We evaluate the \textsc{C-Moral} framework (\textbf{C}ontrollable Multi-\textbf{O}bjective \textbf{M}olecular \textbf{O}ptimization with \textbf{R}einforcement \textbf{A}lignment for \textbf{L}LMs), which integrates customized non-linear property reward design, strict stability constraints, and group-aware advantage aggregation through both GRPO and GDPO. We conduct experiments on two widely adopted open-weight large language models, \textsc{LLaMA} \cite{touvron2023llamaopenefficientfoundation} and \textsc{Mistral} \cite{jiang2023mistral7b}. Our study focuses on the 7B scale to provide a controlled evaluation under a practical and commonly used model size. To isolate the effect of RL post-training, all policy models are initialized from the same SFT checkpoints provided by Dey et al., namely GeLLM$^4$O-C \cite{dey2025large}, and are evaluated under the same benchmark and protocol. The subsequent reinforcement learning post-training is implemented using a custom training pipeline inspired by the efficient design of the veRL framework \cite{Sheng_2025}.

\subsection{Datasets and Tasks}
To evaluate the proposed framework, we use the Controllable Multi-property, Multi-objective Optimization (C-MuMOInstruct) benchmark \cite{dey2025large}. Specifically, we select 10 distinct molecular optimization tasks from the benchmark. To assess performance under different levels of distribution shift, we group these tasks into In-Domain (IND) and Out-of-Domain (OOD) settings, as summarized in Table \ref{tab:dataset_summary}. The IND tasks evaluate the model on property combinations that are closer to the training distribution, while the OOD tasks examine performance on unseen property constraints and scaffold conditions.

For RL post-training, we construct a balanced training set of 100,000 molecules (10,000 per task). The final model is evaluated on a uniform test set of 500 molecules per task.

\begin{table}[t]
\centering
\small
\setlength{\tabcolsep}{4pt} 

\begin{tabular}{@{}cll@{}}
\toprule
\textbf{Type} & \textbf{Task} & \textbf{Target Properties ($\mathcal{P}$-Comb)} \\
\midrule
\multirow{5}{*}{IND} 
& BPQ  & BBBP, PlogP, QED \\
& ELQ  & hERG, LIV, QED \\
& ACEP & AMP, CARC, hERG, PlogP \\
& BDPQ & BBBP, DRD2, PlogP, QED \\
& DHMQ & DRD2, hERG, MUT, QED \\
\cmidrule{1-3}
\multirow{5}{*}{OOD} 
& CDE   & CARC, DRD2, hERG \\
& ABMP  & AMP, BBBP, MUT, PlogP \\
& BCMQ  & BBBP, CARC, MUT, QED \\
& BDEQ  & BBBP, DRD2, hERG, QED \\
& HLMPQ & hERG, LIV, MUT, PlogP, QED \\
\bottomrule
\end{tabular}

\vspace{0.3cm} 

\begin{tabular}{@{}lcc@{\hspace{0.8cm}}lcc@{}}
\toprule
\textbf{Prop.} & $\Delta_p$ & $\Theta_p$ & \textbf{Prop.} & $\Delta_p$ & $\Theta_p$ \\
\midrule
AMP  & 0.1 & 0.8 & HIA   & 0.1 & 0.9 \\
BBBP & 0.1 & 0.8 & LIV   & 0.1 & 0.5 \\
CARC & 0.2 & 0.2 & MUT   & 0.1 & 0.2 \\
DRD2 & 0.1 & 0.4 & PlogP & 1.0 & 1.5 \\
hERG & 0.2 & 0.3 & QED   & 0.1 & 0.9 \\
\bottomrule
\end{tabular}

\caption{Overview of evaluation tasks and property thresholds. \textbf{(Top)} Target property combinations for both IND and OOD scenarios. \textbf{(Bottom)} The target improvement margins ($\Delta_p$) and near-optimal constraints ($\Theta_p$) for the 10 pharmacological properties. }
\label{tab:dataset_summary}
\end{table}

\subsection{Property Calculations}
To provide accurate property feedback during both the reinforcement learning phase and the final evaluation, we employ two established computational oracles. Specifically, we utilize the open-source cheminformatics toolkit RDKit \cite{landrum2013rdkit} to validate generated SMILES strings, extract Morgan fingerprints, and compute fundamental physicochemical properties such as QED and Penalized LogP. For the remaining complex properties, we leverage the ADMET-AI platform \cite{swanson2024admet} as our primary evaluator. These methods have been extensively validated and adopted in recent
studies \cite{averly-etal-2025-liddia, Zholus_Kuznetsov_Schutski_Shayakhmetov_Polykovskiy_Chandar_Zhavoronkov_2025, zheng2025large}.

\subsection{Implementation and Hyperparameters}
We use Low-Rank Adaptation (LoRA) \cite{hu2021loralowrankadaptationlarge} on the projection layers to fine-tune the policy models, rather than updating all parameters. The learning rate is set to $1 \times 10^{-6}$. Both GRPO and GDPO are trained with a batch size of 64 and a group size of 4. Policy updates are performed using 32 mini-batches over 2--3 optimization epochs per rollout.

To preserve valid SMILES syntax during training, we apply a Kullback--Leibler (KL) divergence penalty as a structural regularizer, with a target threshold of 1.0 and adaptive KL coefficients. Full implementation details are provided in Appendix \ref{appendix:implementation} (Table \ref{tab:hyperparams}), and the prompt template is given in Appendix \ref{appendix:prompt}.

\subsection{Baselines and Evaluation Metrics}

We compare our proposed \textsc{C-Moral} framework with two baseline models GeLLM\textsuperscript{4}O-C-P(10)-\text{Mistral} and GeLLM\textsuperscript{4}O-C-P(10)-\textsc{Llama}. For a fair and robust evaluation, all models generate candidate molecules using beam search with a beam width of 20. For each molecule, we select the best candidate according to Algorithm \ref{alg:beam_search}.

To comprehensively assess performance in highly constrained lead optimization, we employ four rigorous metrics: \textbf{(1) Success Optimized Rate (\textsc{Sor})}: the proportion of molecules that improve the targeted properties while maintaining other stability constraints; \textbf{(2) Strict Success Optimized Rate (\textsc{Ssor})}:  percentage of candidates that improve sub-optimal properties and strictly maintain near-optimal ones while preserving the core scaffold; \textbf{(3) Similarity (\textsc{Sim})}: the Tanimoto similarity over Morgan fingerprints between generated candidates and initial molecules; \textbf{(4) Relative Improvement (\textsc{Ri})}: the relative improvement across all sub-optimal properties. The detailed implementation is in Appendix \ref{appendix:metrics}. 

\section{Results}
\label{sec:results}

In this section, we present a comprehensive evaluation of the proposed \textsc{C-Moral} framework on the C-MuMOInstruct benchmark. Our analysis is structured as follows: we first examine the main performance gains across both In-Domain (IND) and Out-of-Domain (OOD) tasks to illustrate how \textsc{C-Moral} achieves better Pareto trade-offs without sacrificing scaffold integrity; next, we analyze the impact of our group-relative alignment and non-linear reward shaping in preventing implicit property violations; and finally, we compare our results against state-of-the-art baselines for 7B-scale molecular language models.

\subsection{Ablation Analysis}
To systematically validate the structural design of our continuous multi-objective reward formulation, we conducted an ablation study on the In-Domain (IND) task (Table \ref{tab:ablation_ind}) based on the \textsc{Mistral} model. We dissect the contributions of our two core mechanisms: Property Score Sigmoid Alignment and Non-linear Reward Aggregation.

\begin{table}[ht]
\centering
\caption{Ablation study on the IND task using \textsc{Mistral}. Non-linear aggregation prevents reward collapse (recovers Sim), while Sigmoid alignment maximizes overall success rates. Best and second-best results are bold and underlined.}
\label{tab:ablation_ind}
\footnotesize 
\setlength{\tabcolsep}{4pt} 
\renewcommand{\arraystretch}{1.2}
\resizebox{\columnwidth}{!}{
\begin{tabular}{@{} l cccc @{}}
\toprule
\multirow{2}{*}{Method} & \multicolumn{4}{c}{IND Task} \\
\cmidrule(l){2-5}
 & \textbf{SOR(\%)}$\uparrow$ & \textbf{SSOR(\%)}$\uparrow$ & \textbf{Sim}$\uparrow$ & \textbf{RI}$\uparrow$ \\
\midrule
GeLLM$^4$O-C$_{\mathrm{Mistral}}$ & 33.7 & 14.4 & \underline{0.58} & 51.4 \\
\midrule
GRPO w/ Linear AM & 24.2 & 11.2 & 0.35 & 21.5 \\
\quad + Geometric Mean & 39.7 & 17.1 & \underline{0.58} & \underline{108.4} \\
\quad \quad + Sigmoid Align & \textbf{48.9} & \textbf{25.1} & \textbf{0.59} & 96.1 \\
\midrule
GDPO w/ Linear AM & 9.5 & 4.5 & 0.28 & 12.3 \\
\quad + LogSum-Exp & 37.3 & 15.6 & 0.57 & 104.2 \\
\quad \quad  + Sigmoid Align & \underline{47.0} & \underline{25.0} & \textbf{0.59} & \textbf{110.4} \\
\bottomrule
\end{tabular}
}
\end{table}

\begin{figure*}[t] 
    \centering
    \includegraphics[width=\textwidth]{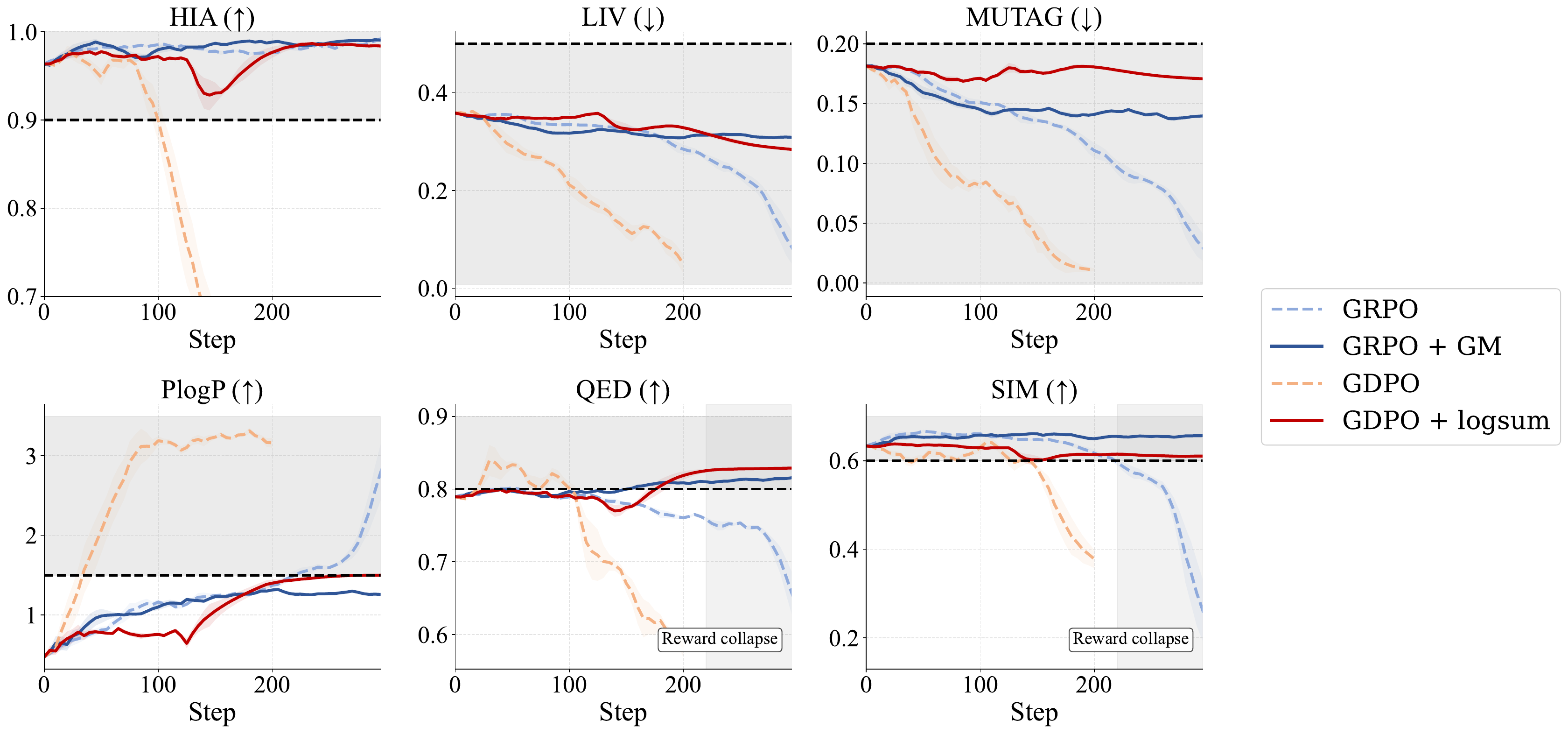}
    \caption{Ablation study of reward aggregation on the HLMPQ task using \textsc{Mistral}-7B over 300 training steps. The results demonstrate that linear Arithmetic Mean (AM) methods suffer from varying degrees of reward collapse. In contrast, our non-linear approaches successfully maintain all property values within their safe boundaries.} 
    \label{fig:ablation_aggregation}
\end{figure*}

\paragraph{Vulnerability of Linear AM and Reward Collapse.}
The aggregation function critically affects how the policy balances multiple objectives. As shown in Table \ref{tab:ablation_ind}, linear Arithmetic Mean (AM) causes clear degradation relative to the SFT baseline for both GRPO and GDPO. Under GRPO, SOR drops from $33.7\% \to 24.2\%$ and Sim from $0.58 \to 0.35$. The effect is even more severe for GDPO, where SOR further falls to $9.5\%$ and Sim to $0.28$. These results suggest that linear scalarization tends to over-optimize a subset of easy objectives while sacrificing others, leading to unbalanced optimization and poor scaffold preservation. This trend is further supported by Figure \ref{fig:ablation_aggregation}, where AM-based training fails to keep all properties within their safe regions on the more challenging HLMPQ task.

\paragraph{Effectiveness of Non-Linear Reward Aggregation.}
Replacing AM with non-linear aggregation substantially improves optimization balance. Under GRPO, switching from linear AM to Geometric Mean improves SOR from $24.2\% \to 39.7\%$ (+64.0\% relative) and restores Sim from $0.35 \to 0.58$, while SSOR also increases from $11.2\% \to 17.1\%$. A similar trend is observed for GDPO: replacing AM with LogSum-Exp raises SOR from $9.5\% \to 37.3\%$ and Sim from $0.28 \to 0.57$, with SSOR improving from $4.5\% \to 15.6\%$. These recovered Sim values are close to the SFT baseline, indicating that non-linear aggregation effectively prevents the implicit sacrifice of scaffold fidelity. Figure \ref{fig:ablation_aggregation} further shows that these non-linear designs produce much more stable training trajectories across all properties. In contrast to the instability and reward collapse of GRPO and GDPO, the non-linear variants maintain smooth improvements and achieve stronger final performance.

\paragraph{Effectiveness of Property Score Sigmoid Alignment.}
Building upon non-linear aggregation, Property Score Sigmoid Alignment maps heterogeneous objectives into a unified $[0, 1]$ scale, directly addressing their inherent scale mismatch. The empirical gains are substantial: for GRPO, adding Sigmoid Align to Geometric Mean increases SOR from $39.7\% \to 48.9\%$ (+23.2\% relative). Crucially, it boosts the strict success rate (SSOR) from $17.1\% \to 25.1\%$, indicating finer control over all constraints simultaneously, while maintaining strong scaffold similarity ($0.58 \to 0.59$). For GDPO, adding it to LogSum-Exp improves SOR from $37.3\% \to 47.0\%$ and SSOR from $15.6\% \to 25.0\%$, with Sim slightly increasing ($0.57 \to 0.59$). These results suggest that, beyond preventing reward collapse, sigmoid alignment further improves optimization efficiency. By reducing scale mismatch, it prevents metrics with broader ranges from dominating and shifts learning toward the remaining property bottlenecks.

\begin{table*}[t]
\centering
\begin{minipage}[t]{0.65\textwidth}
\vspace{0pt}
\centering
(a)\par\vspace{2pt}
\small
\setlength{\tabcolsep}{4pt}
\renewcommand{\arraystretch}{1.1}
\resizebox{\textwidth}{!}{
\begin{tabular}{@{} l cccc cccc @{}}
\toprule
\multirow{2}{*}{\textbf{Model}} & \multicolumn{4}{c}{In-Domain (IND)} & \multicolumn{4}{c}{Out-of-Domain (OOD)} \\
\cmidrule(lr){2-5} \cmidrule(lr){6-9}
& \textbf{SOR}$^{\uparrow}$ & \textbf{SSOR}$^{\uparrow}$ & \textbf{Sim}$^{\uparrow}$ & \textbf{RI}$^{\uparrow}$ & \textbf{SOR}$^{\uparrow}$ & \textbf{SSOR}$^{\uparrow}$ & \textbf{Sim}$^{\uparrow}$ & \textbf{RI}$^{\uparrow}$ \\
\midrule

\rowcolor{gray!10}[0pt][0pt] \multicolumn{9}{@{}l@{}}{\textit{Supervised Fine-Tuning (SFT) Baselines}} \\
\quad GeLLM$^4$O-C$_{\mathrm{Mistral}}$ & 33.7 & 14.4 & \underline{0.58} & 51.4 & 24.4 & 9.8 & \textbf{0.60} & \textbf{27.3} \\
\quad GeLLM$^4$O-C$_{\mathrm{Llama}}$   & 30.2 & 13.7 & 0.56 & 61.3 & 25.2 & 10.2 & 0.57 & 6.5 \\
\midrule

\rowcolor{gray!10}[0pt][0pt] \multicolumn{9}{@{}l@{}}{\textit{RL Post-Training (Ours: \textsc{C-Moral})}} \\
\quad GRPO$_{\mathrm{Mistral}}$ & \textbf{48.9} & \underline{25.1} & \textbf{0.59} & 96.1 
                   & \textbf{39.5} & \textbf{20.8} & \textbf{0.60} & 19.5 \\
\quad GDPO$_{\mathrm{Mistral}}$ & \underline{47.0} & 25.0 & \textbf{0.59} & \textbf{110.4} & \underline{38.3} & \underline{19.8} & \underline{0.59} & \underline{27.2} \\
\quad GRPO$_{\mathrm{Llama}}$   & 40.3 & 20.1 & \underline{0.58} & 102.6 & 37.2 & 19.6 & \textbf{0.60} & 10.5 \\
\quad GDPO$_{\mathrm{Llama}}$   & 43.4 & \textbf{25.4} & 0.57 & \underline{110.3} & 35.6 & 19.5 & \underline{0.59} & 19.1 \\
\bottomrule
\end{tabular}
}
\end{minipage}
\hfill
\begin{minipage}[t]{0.34\textwidth}
\vspace{0pt}
\centering
(b)\par\vspace{2pt}
\includegraphics[width=\textwidth,height=0.35\textheight,keepaspectratio]{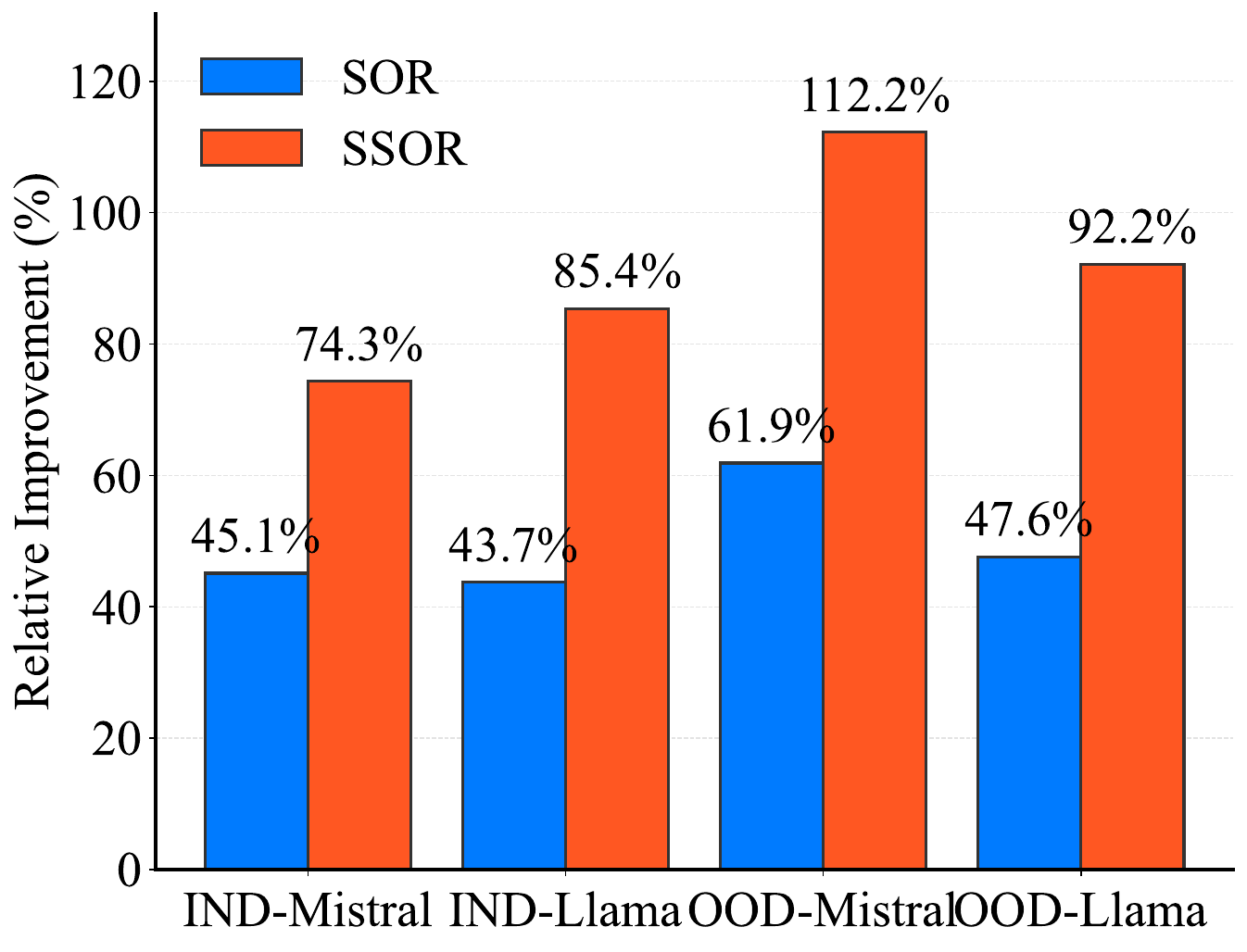}
\end{minipage}

\caption{Overall results on In-Domain (IND) and Out-of-Domain (OOD) tasks. (a) Average performance comparison between baselines and \textsc{C-Moral} variants in SOR, SSOR, similarity, and RI. (b) Relative improvements in SOR and SSOR over the baselines. Bold denotes the best result in each column, and underlined denotes the second-best.}
\label{tab:overall_summary}
\end{table*}

\subsection{Comparison with Baselines}
\label{subsec:main_results}

We next evaluate our proposed \textsc{C-Moral} framework on \textsc{Mistral} and \textsc{Llama} architectures and compare it against strong SFT baselines (GeLLM$^4$O-C). Table \ref{tab:overall_summary} presents the overall average results on both In-Domain (IND) and Out-of-Domain (OOD) tasks, including both our GRPO and GDPO post-training variants. 

\paragraph{Performance on In-Domain (IND) Tasks.}
On IND tasks, RL post-training consistently improves over the SFT baselines: (1) For \textsc{Mistral} base, GRPO improves SOR from $33.7\% \to 48.9\%$ (+45.1\%) and SSOR from $14.4\% \to 25.1\%$ (+74.3\%), while maintaining high similarity. GDPO achieves comparable SOR ($33.7\% \to 47.0\%$, +39.5\%) and SSOR ($14.4\% \to 25.0\%$, +73.6\%), while substantially improving RI ($51.4 \to 110.4$, +114.8\%); (2) For \textsc{Llama}, GRPO improves SOR from $30.2\% \to 40.3\%$ (+33.4\%) and SSOR from $13.7\% \to 20.1\%$ (+46.7\%), with Sim increasing from $0.56 \to 0.58$. GDPO further improves SOR to $43.4\%$ (+43.7\%) and SSOR to $25.4\%$ (+85.4\%), while raising RI from $61.3 \to 110.3$ (+79.9\%). Detailed IND results are provided in Table~\ref{tab:ind_results}. Overall, these results indicate that \textsc{C-Moral} consistently strengthens in-domain optimization performance, while GDPO shows a slight advantage on stricter metrics and RI.

\paragraph{Generalization to Out-of-Domain (OOD) Tasks.}
OOD tasks are more challenging for all models, and the SFT baselines show clear drops from IND to OOD. RL post-training remains consistently effective. For \textsc{Mistral}, GRPO improves OOD SOR from $24.4\% \to 39.5\%$ (+61.9\%) and SSOR from $9.8\% \to 20.8\%$ (+112.2\%), while keeping Sim unchanged at 0.60. GDPO also performs strongly, improving SOR from $24.4\% \to 38.3\%$ (+57.0\%) and SSOR from $9.8\% \to 19.8\%$ (+102.0\%). For \textsc{Llama}, GRPO improves SOR from $25.2\% \to 37.2\%$ (+47.6\%) and SSOR from $10.2\% \to 19.6\%$ (+92.2\%), while increasing Sim from $0.57 \to 0.60$. GDPO improves SOR from $25.2\% \to 35.6\%$ (+41.3\%), SSOR from $10.2\% \to 19.5\%$ (+91.2\%), and RI from $6.5 \to 19.1$ (+193.8\%). Detailed OOD results are provided in Table~\ref{tab:ood_results}.

\paragraph{Similarity Comparison.}
Both RL variants generally improve similarity over the corresponding SFT baselines. On IND tasks, \textsc{Mistral} improves from $0.58 \to 0.59$, while \textsc{Llama}-GRPO improves from $0.56\to0.58$. On OOD tasks, \textsc{Mistral} remains at 0.59\textasciitilde0.60, and \textsc{Llama} improves from $0.57 \to 0.60$ with GRPO and to 0.59 with GDPO. The average similarity still does not always reach our target of 0.60, partly because beam search (\texttt{num\_beam}) favors better-scoring candidates at the cost of slightly larger structural edits.

\paragraph{Overall observations.}
Overall, \textsc{C-Moral} consistently outperforms the SFT baselines across both backbones and both settings. The best IND SOR improves from $33.7\%/30.2\%$ to $48.9\%/43.4\%$ for \textsc{Mistral}/\textsc{Llama}, corresponding to relative gains of +45.1\% and +43.7\%, respectively; the best IND SSOR improves from $14.4\%/13.7\%$ to $25.1\%/25.4\%$, yielding +74.3\% and +85.4\% relative improvements. On OOD tasks, the best SOR improves from $24.4\%/25.2\%$ to $39.5\%/37.2\%$ (+61.9\% / +47.6\%), and the best SSOR improves from $9.8\%/10.2\%$ to $20.8\%/19.6\%$ (+112.2\% / +92.2\%). These gains are achieved while largely preserving scaffold similarity, indicating that the improvements come from better optimization rather than structural drift.

\section{Conclusion}
We introduced \textsc{C-Moral}, a reinforcement learning post-training framework for controllable multi-objective molecular optimization. The proposed framework is built on three key contributions: (1) group-based relative optimization, instantiated through memory-efficient GRPO and GDPO variants, to support stable policy learning under diverse optimization dynamics; (2) property score sigmoid alignment for handling heterogeneous objective scales and enabling fine-grained controllability; and (3) continuous non-linear reward aggregation, which enforces balanced multi-objective trade-offs and effectively prevents reward collapse.

Empirical results on the C-MuMO benchmark demonstrate that \textsc{C-Moral} substantially improves both success rate and strict success rate over strong SFT baselines while maintaining scaffold similarity. Furthermore, these performance improvements seamlessly extend from in-domain tasks to significantly more challenging out-of-domain settings. This robust generalization capability suggests that our RL post-training approach effectively navigates unseen chemical spaces under distribution shifts. Ultimately, \textsc{C-Moral} highlights the promising potential of aligning language models with complex pharmacological objectives, offering a scalable and reliable tool to optimize molecules.


\section*{Limitations}
Although our framework may extend beyond the benchmark studied here, the empirical gains reported in this work are still specific to the molecular optimization settings covered by C-MuMO. Our current study has several limitations. (1) We mainly focus on improving performance within an existing benchmark, and do not yet validate the proposed framework on a broader range of datasets or molecular optimization tasks beyond C-MuMO. While the OOD results suggest promising transferability, further experiments are still needed to assess its effectiveness in both single-objective and multi-objective settings under more diverse task formulations. (2) Our experiments are limited to two 7B-scale backbone models, \textsc{Mistral} and \textsc{Llama}, and we do not explore a wider range of model families, model scales, or alternative post-training strategies such as chain-of-thought-style optimization. As a result, it remains unclear how broadly the proposed framework generalizes across architectures and optimization paradigms. (3) Our evaluation relies on widely used molecular property metrics and predictors, which may still introduce inaccuracies when estimating true molecular quality. In addition, this work does not aim to train a single highly generalizable model that can cover a wide range of molecular optimization scenarios. Future work should therefore evaluate \textsc{C-Moral} on more diverse datasets and tasks, improve the reliability of molecular evaluation, and investigate how to build more generalizable molecular optimization models.

\section*{Ethics Statement}
Our work, C-MORAL, focuses on accelerating computational drug design. While molecular generation models inherently carry dual-use risks (such as being misused to generate toxic compounds), our research strictly evaluates on benign, standard pharmacological benchmarks (C-MuMOInstruct). We emphasize that C-MORAL is an in silico tool designed to assist medicinal chemists. Any generated molecules are purely computational predictions and require rigorous laboratory synthesis and safety validation before any real-world application.

\section*{Acknowledgements}

This work used the Delta system at the National Center for Supercomputing Applications (NCSA) through allocation CIS251097 from the Advanced Cyberinfrastructure Coordination Ecosystem: Services \& Support (ACCESS) program \cite{boerner2023access}. We gratefully acknowledge the computing resources provided by NCSA and the ACCESS program.

\clearpage
\bibliography{custom}

\clearpage
\appendix

\section{Details on Reward Aggregation}
\label{sec:appendix}

In multi-objective molecular optimization, the choice of reward aggregation directly determines the alignment behavior of the RL agent. We deliberately avoid standard linear scalarization in favor of non-linear aggregation methods, such as the Geometric Mean (GM) or the Smooth Minimum (-Log-Sum-Exp). 

\subsection{Reward Aggregation}
\label{appendix:aggregation}
The standard Arithmetic Mean (AM) defines the total reward as a linear combination of individual objectives: $R_{AM} = \frac{1}{N} \sum_{i=1}^N r_i$. While mathematically simple, it exhibits severe limitations when navigating complex chemical spaces. 

\begin{figure}[t]
  \centering
  \includegraphics[width=\columnwidth]{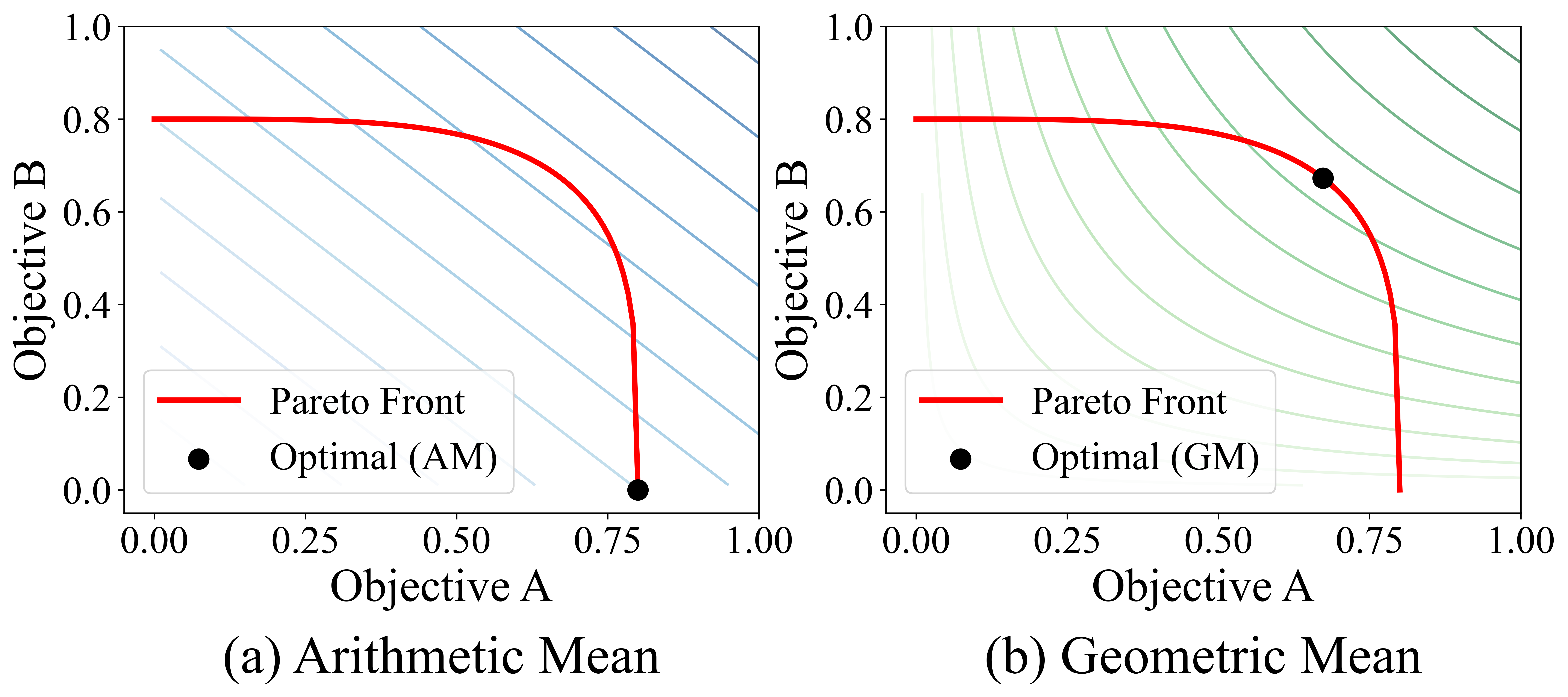}
  \caption{Optimal solutions on a concave Pareto front. (a) Arithmetic Mean leads to extreme boundary solutions. (b) Geometric Mean forces balanced trade-offs.}
  \label{fig:pareto_front}
\end{figure}

As illustrated in Figure \ref{fig:pareto_front}a, the level curves of AM are straight lines. In molecular optimization, the true Pareto front between competing pharmacological properties (e.g., binding affinity versus drug-likeness) is typically strictly concave. Optimizing a linear objective over a concave front inevitably drives the solution toward the extremes (corner solutions). This mathematical artifact is the root cause of the "implicit sacrifice" phenomenon, where the LLM maximizes one easily optimizable property by completely violating another.

In contrast, Figure \ref{fig:pareto_front}b demonstrates the behavior of the Geometric Mean: 
\[
R_{GM} = \left( \prod_{i=1}^N r_i \right)^{\frac{1}{N}}. 
\]
The GM constructs strictly convex, hyperbolic level sets. When these level sets intersect with the concave Pareto front, the optimal solution naturally settles near the center. This geometry enforces balanced trade-offs, effectively acting as a differentiable logical "AND" operator that requires all constraints to be reasonably satisfied.

\subsection{Reward Aggregation: GRPO \& GDPO}
\label{appendix:gm_vs_logmin}

To enforce the simultaneous satisfaction of all constraints, the aggregation function must approximate the logical AND ($\min$) operator. While both Geometric Mean (GM) and -Log-Sum-Exp (LogMin) provide continuous gradients, their mathematical properties dictate their specific applications in GRPO and GDPO.

\begin{figure}[t]
  \centering
  \includegraphics[width=\columnwidth]{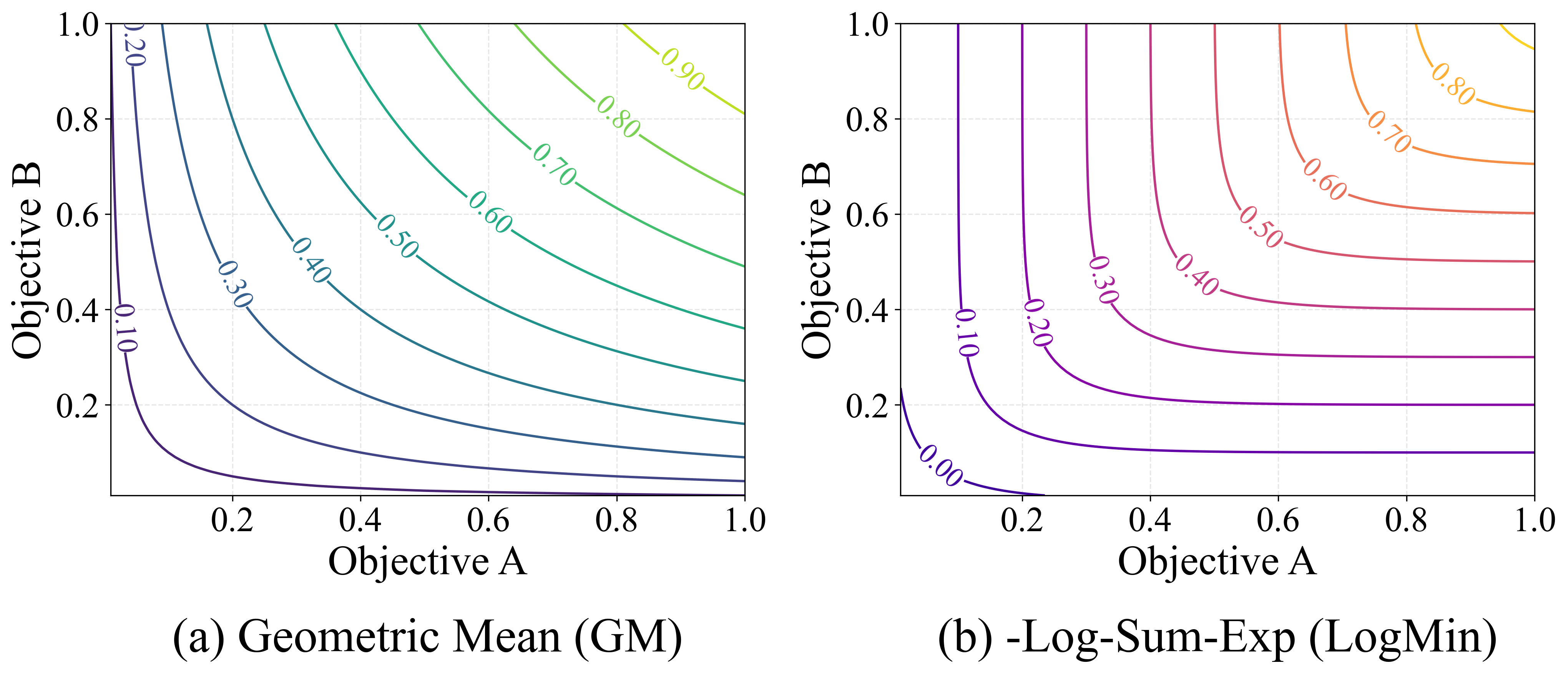}
  \caption{Contour plots of aggregation functions on 2-dimension objectives. (a) Geometric Mean (b) LogSum}
  \label{fig:contour}
\end{figure}

As shown in Figure \ref{fig:contour}, GM and LogMin exhibit fundamentally different gradient behaviors. 

For GM, defined as $R_{GM} = (\prod_{i=1}^N r_i)^{\frac{1}{N}}$, the gradient with respect to a single objective $r_j$ is:
$$ \frac{\partial R_{GM}}{\partial r_j} = \frac{1}{N} \frac{R_{GM}}{r_j} $$
This multiplicative nature ensures that the optimization of $r_j$ is holistically scaled by the performance of all other objectives.

Conversely, LogMin acts as a strict bottleneck. Defined as $R_{LSE} = -\frac{1}{k} \log \sum_{i=1}^N \exp(-k \cdot x_i)$, its gradient is exactly the softmax distribution:
$$ \frac{\partial R_{LSE}}{\partial x_j} = \frac{\exp(-k \cdot x_j)}{\sum_{i=1}^N \exp(-k \cdot x_i)} $$
When $k$ is sufficiently large, if $x_j$ is the worst-performing objective, its exponential term dominates the denominator. Consequently, $\frac{\partial R_{LSE}}{\partial x_j} \to 1$, while gradients for all other objectives approach $0$. This mathematically explains the "L-shaped" contours in Figure \ref{fig:contour}(b), proving that LogMin forces the model to exclusively penalize the weakest property.

\paragraph{Domain Constraints}
Beyond gradients, the mathematical domain strictly determines algorithmic compatibility:
\begin{itemize}
    \item \textbf{GRPO ($r_i \in [0, 1]$):} GRPO maps raw properties to positive scores via non-linear reward shaping, perfectly satisfying the strict non-negativity requirement of the GM function.
    \item \textbf{GDPO ($A_i \in \mathbb{R}$):} GDPO optimizes pairwise advantages, which naturally span negative values. Since GM is undefined for negative numbers, LogMin becomes mathematically mandatory. It seamlessly aggregates unbounded real numbers, isolating the most critical negative margin without breaking domain constraints.
\end{itemize}

\begin{figure*}[t] 
  \centering
  \includegraphics[width=\textwidth]{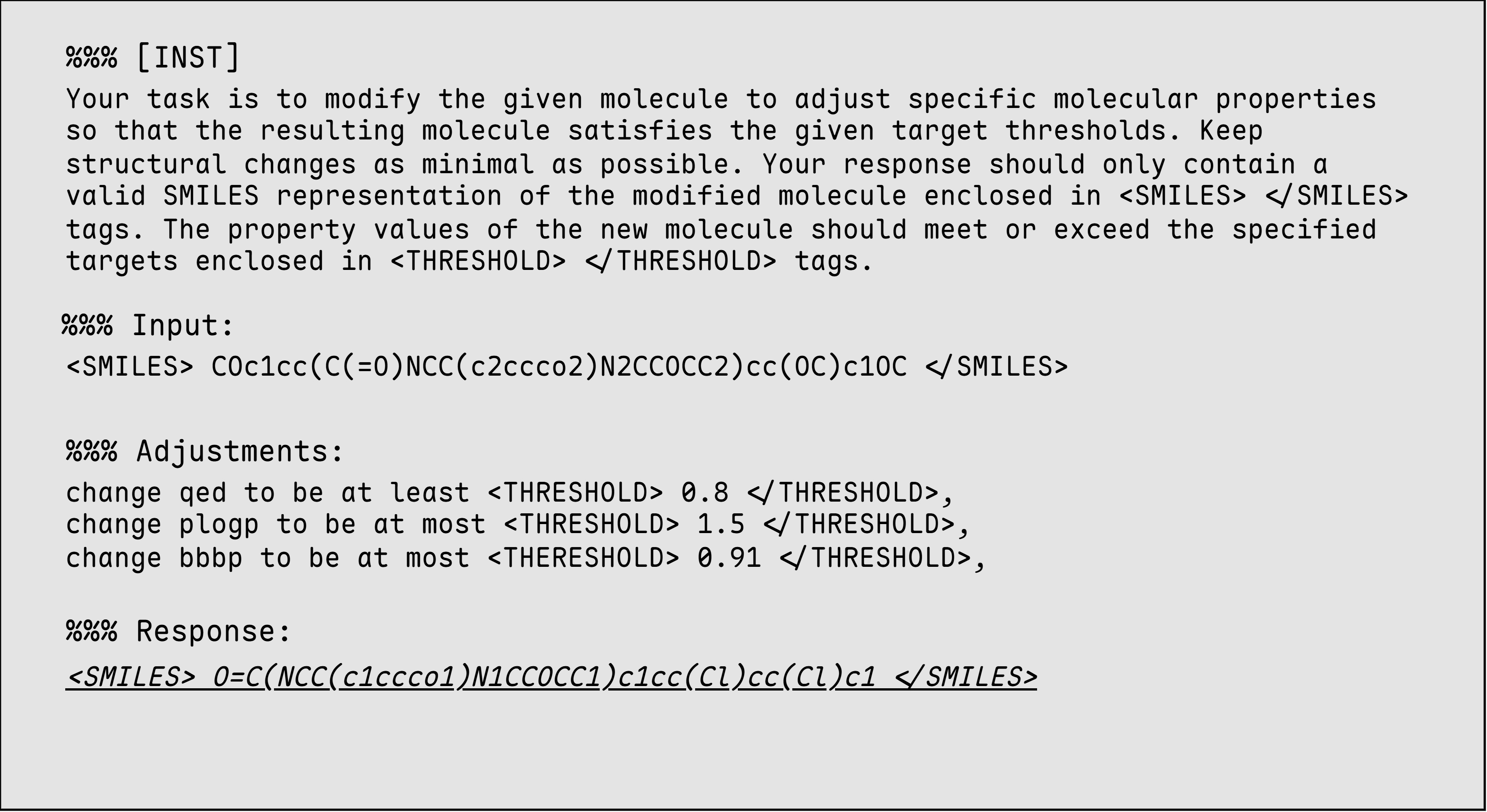} 
  \caption{An example of the highly structured prompt template used in \textsc{C-Moral}. The prompt dynamically integrates the source molecule (Input) with explicit optimization directions and numerical thresholds (Adjustments) to guide the language model during the RL phase.}
  \label{fig:prompt}
\end{figure*}

\section{Details on Implementation}
\label{appendix:implementation}
\subsection{Hyperparameters}
\label{appendix:hyperparams}

\begin{table}[ht]

\centering
\small
\begin{tabular}{ll}
\toprule
\textbf{Hyperparameter} & \textbf{Value} \\
\midrule
Base Model & [Mistral-7B-v0.3, Llama3.1-8B] \\
Optimizer & AdamW ($\beta_1=0.9, \beta_2=0.95$) \\
Learning Rate & $1 \times 10^{-6}$ \\
Learning Rate Scheduler & Cosine with 10\% warmup \\
LoRA Rank ($r$) & 16 \\
LoRA Alpha ($\alpha$) & 32 \\
Target Modules & \makecell[l]{q\_proj, v\_proj, k\_proj, o\_proj,\\ gate, up, down\_proj, lm\_head} \\
Max Sequence Length & 100 tokens \\
Temperature & 1.0 \\
\midrule
RL Algorithm & GRPO/GDPO \\
Group Size ($G$) & 4 \\
KL Coefficient (Initial) & 0.05 \\
KL Target Value & 1.0 \\
Rollout Batch Size & 32 \\
Optimization Epochs & 2 \\
MiniBatch Size & 32 \\
Training Size & 10000 mols per task \\ 
\midrule
Hardware & 1 $\times$ NVIDIA A100 (80GB) \\
Training Time & $\sim$6 Hours per task \\
\bottomrule
\end{tabular}
\caption{Hyperparameters for SFT and RL post-training stages.}
\label{tab:hyperparams}
\end{table}

\begin{algorithm*}[t]
\caption{Candidate Selection}
\label{alg:beam_search}
\begin{algorithmic}[1]
\Require Source molecule $M_{src}$, trained policy $\pi_\Theta$, beam size $K=20$
\Require Target properties $\mathcal{P}$, near-optimal thresholds $\Theta_p$ for $p \in \mathcal{P}$
\Require Optimization directions $d_p \in \{1, -1\}$ for $p \in \mathcal{P}$ \Comment{$1$: Maximize, $-1$: Minimize}
\Require Target improvement margins $\Delta_p$, and stability tolerance margins $\Delta_q$
\Ensure The best aligned candidate $M_{best}$

\State \textbf{Step 1:} \textit{Evaluate source molecule and dynamically formulate optimization tasks}
\State $\mathcal{P}_{sub} \leftarrow \{ p \in \mathcal{P} \mid d_p \cdot p(M_{src}) < d_p \cdot \Theta_p \}$ \Comment{Identify sub-optimal properties}
\State $\mathcal{P}_{con} \leftarrow \{ q \in \mathcal{P} \mid d_p \cdot q(M_{src}) \ge d_p \cdot \Theta_q \}$ \Comment{Identify near-optimal properties}
\State

\State \textbf{Step 2:} \textit{Generate candidates conditioned on the formulated tasks}
\State $\mathcal{M}_{gen} \leftarrow \text{BeamSearch}(\pi_\Theta(\cdot \mid M_{src}), K)$ 
\State

\State \textbf{Step 3:} \textit{Filter candidates to construct the SOR-compliant set $\mathcal{C}_{SOR}$}
\State $\mathcal{C}_{SOR} \leftarrow \left\{ c \in \mathcal{M}_{gen} \ \middle| \ 
    \begin{array}{l}
        \forall p \in \mathcal{P}_{sub}: d_p \cdot (p(c) - p(M_{src})) \ge \Delta_p \ \land \\
        \forall q \in \mathcal{P}_{con}: |q(c) - q(M_{src})| \le \Delta_q
    \end{array}
\right\}$
\State

\State \textbf{Step 4:} \textit{Select the optimal candidate based on Relative Improvement (RI)}
\If{$\mathcal{C}_{SOR} \neq \emptyset$}
    \State $M_{best} \leftarrow \underset{c \in \mathcal{C}_{SOR}}{\arg\max} \ \text{RI}(c, M_{src})$
\Else
    \State $M_{best} \leftarrow \underset{c \in \mathcal{M}_{gen}}{\arg\max} \ \text{RI}(c, M_{src})$ \Comment{Fallback: maximum RI relaxation}
\EndIf

\State \Return $M_{best}$
\end{algorithmic}
\end{algorithm*}
\subsection{Prompt Design}
\label{appendix:prompt}

To ensure the language model accurately interprets the multi-objective optimization tasks, we design a highly structured and dynamically assembled prompt template, as illustrated in Figure \ref{fig:prompt}. Instead of using static instructions, the assembly process programmatically injects task-specific information into the template for each generated episode. Specifically, the \texttt{Input} field is dynamically substituted with the exact SMILES string of the sampled source molecule. Furthermore, the \texttt{Adjustments} field acts as a condition-aware directive that translates mathematical constraints into natural language. It is procedurally generated by evaluating the source molecule's initial property values against the predefined near-optimal thresholds ($\Theta_p$). If a target property is sub-optimal, the prompt formulates a clear objective to improve it (e.g., specifying "to be at least" a certain threshold). Conversely, if the property already exceeds $\Theta_p$, the prompt imposes a strict maintenance constraint to prevent implicit sacrifice. By explicitly wrapping these quantitative targets in \texttt{<THRESHOLD>} tags, we compel the model to align its structural modifications directly with the desired multi-objective boundaries, ultimately producing a valid candidate enclosed in \texttt{<SMILES>} tags without generating superfluous text.

\section{Details on Evaluation Metrics}
\label{appendix:metrics}

To rigorously evaluate the performance of lead optimization, we formulate our four evaluation metrics mathematically. Let $M_{src}$ denote the initial hit (source) molecule and $M_{gen}$ denote the generated candidate. We partition the pharmacological properties into two sets: $\mathcal{P}_{sub}$ representing the targeted sub-optimal properties that require improvement, and $\mathcal{P}_{con}$ representing the stability constraints or near-optimal properties that must be maintained.

\paragraph{1. Similarity (\textsc{Sim})}
To quantify structural preservation, we compute the Tanimoto similarity over Morgan fingerprints (radius 2, 2048 bits) between the generated candidate and the initial hit. All fingerprint generations and similarity calculations are implemented using the open-source cheminformatics RDKit. This ensures that the optimization strictly occurs within the valid chemical neighborhood of the lead compound. 
\[
\begin{aligned}
    \textsc{Sim}(M_{src}, M_{gen}) = & \\
    \frac{|\text{Morgan}(M_{src}) \cap \text{Morgan}(M_{gen})|}{|\text{Morgan}(M_{src}) \cup \text{Morgan}(M_{gen})|}
\end{aligned}
\]
The overall \textsc{Sim} score reported in our results is the average Tanimoto similarity across all valid generated molecules.

\paragraph{2. Success Optimized Rate (\textsc{Sor})}
To ensure that the reported improvements are chemically meaningful and robust to model noise, we introduce a significance margin $\Delta_k$. We divide the target properties into two subsets: $\mathcal{P}_{sub}$, containing properties that require directional optimization, and $\mathcal{P}_{con}$, containing properties that should remain stable relative to the source hit.

For each property $p \in \mathcal{P}_{sub}$, we define a direction indicator $s_p \in \{+1,-1\}$, where $s_p=+1$ denotes that the property is expected to increase and $s_p=-1$ denotes that it is expected to decrease. A generated molecule is considered successful on the optimization subset if all such properties improve beyond the significance margin:
\[
    C_{sub}^{(i)} = \prod_{p \in \mathcal{P}_{sub}} \mathbb{I}\left( s_p \bigl(p(M_{gen}^{(i)}) - p(M_{src}^{(i)})\bigr) \geq \Delta_k \right).
\]

For each property $q \in \mathcal{P}_{con}$, we require the generated molecule to remain within a tolerance band around the source hit:
\[
    C_{con}^{(i)} = \prod_{q \in \mathcal{P}_{con}} \mathbb{I}\left( \left| q(M_{gen}^{(i)}) - q(M_{src}^{(i)}) \right| \le \Delta_k \right).
\]

The overall \textsc{Sor} is then defined as the fraction of hit-candidate pairs that satisfy both conditions:
\[
    \textsc{Sor} = \frac{1}{N} \sum_{i=1}^N \left( C_{sub}^{(i)} \cdot C_{con}^{(i)} \right).
\]

In our experiments, $\Delta_k$ is chosen according to the standard deviation or the known error bar of the corresponding property predictor.

\paragraph{3. Strict Success Optimized Rate (\textsc{Ssor})}
While \textsc{Sor} evaluates whether a generated molecule achieves meaningful directional improvements on the target subset while preserving the constrained subset within tolerance, it does not require the final molecule to satisfy strict near-optimal criteria on all relevant properties. We therefore introduce the \textsc{Strict Success Optimized Rate} (\textsc{Ssor}), which measures the fraction of generated molecules whose properties all fall within the desired near-optimal region.

Let $\Theta_r$ denote the near-optimal threshold for property $r \in \mathcal{P}$, where the satisfaction direction depends on the property type. We define the strict success indicator for the $i$-th pair as
\[
    C_{strict}^{(i)} = \prod_{r \in \mathcal{P}} \mathbb{I}\bigl( \mathrm{sat}_r(M_{gen}^{(i)}; \Theta_r) = 1 \bigr),
\]
where $\mathrm{sat}_r(\cdot;\Theta_r)$ is a property-specific satisfaction function indicating whether property $r$ meets its corresponding near-optimal threshold.

The overall \textsc{Ssor} is then defined as
\[
    \textsc{Ssor} = \frac{1}{N} \sum_{i=1}^N C_{strict}^{(i)}.
\]

In other words, \textsc{Ssor} is a stricter metric than \textsc{Sor}, requiring all relevant properties to reach their prescribed near-optimal targets simultaneously.

\paragraph{4. Relative Improvement (\textsc{Ri})}
While \textsc{Sor} and \textsc{Ssor} measure whether a generated molecule satisfies the desired optimization criteria, they do not reflect the magnitude of improvement. We therefore define the \textsc{Relative Improvement} (\textsc{Ri}) to quantify the average directional relative change on the subset of sub-optimal properties.

For each property $p \in \mathcal{P}_{sub}$, let $s_p \in \{+1,-1\}$ denote its desired optimization direction, where $s_p=+1$ indicates that the property is expected to increase and $s_p=-1$ indicates that it is expected to decrease. We first define the instance-level relative improvement as
\[
\textsc{Ri}^{(i)}
=
\frac{1}{|\mathcal{P}_{sub}|}
\sum_{p \in \mathcal{P}_{sub}}
\frac{
s_p \bigl(p(M_{gen}^{(i)}) - p(M_{src}^{(i)})\bigr)
}{
\left|p(M_{src}^{(i)})\right|
}.
\]

The overall \textsc{Ri} is then computed by averaging over all molecule pairs:
\[
\textsc{Ri}
=
\frac{1}{N}\sum_{i=1}^{N}\textsc{Ri}^{(i)}.
\]

In this way, \textsc{Ri} measures the average relative change on sub-optimal properties in their desired optimization directions: positive values indicate improvement, while negative values indicate movement against the target direction.

\begin{table*}[t]
\centering
\footnotesize
\setlength{\tabcolsep}{1.8pt}
\renewcommand{\arraystretch}{1.22}
\resizebox{\textwidth}{!}{
\begin{tabular}{@{}l ccc@{\hspace{4pt}} ccc@{\hspace{4pt}} ccc@{\hspace{4pt}} ccc@{\hspace{4pt}} ccc@{}}
\toprule
\multirow{2}{*}{\textbf{Model}} & \multicolumn{3}{c}{\textsc{Bpq}} & \multicolumn{3}{c}{\textsc{Elq}} & \multicolumn{3}{c}{\textsc{Acep}} & \multicolumn{3}{c}{\textsc{Bdpq}} & \multicolumn{3}{c}{\textsc{Dhmq}} \\
\cmidrule(lr){2-4} \cmidrule(lr){5-7} \cmidrule(lr){8-10} \cmidrule(lr){11-13} \cmidrule(lr){14-16}
& S/SS$^{\uparrow}$ & Sim$^{\uparrow}$ & RI$^{\uparrow}$ & S/SS$^{\uparrow}$ & Sim$^{\uparrow}$ & RI$^{\uparrow}$ & S/SS$^{\uparrow}$ & Sim$^{\uparrow}$ & RI$^{\uparrow}$ & S/SS$^{\uparrow}$ & Sim$^{\uparrow}$ & RI$^{\uparrow}$ & S/SS$^{\uparrow}$ & Sim$^{\uparrow}$ & RI$^{\uparrow}$ \\
\midrule

\rowcolor{gray!15} \multicolumn{16}{c}{\textit{Supervised Fine-Tuning (SFT) Baselines}} \\
GeLLM$^4$O-C$_{\mathrm{Mistral}}$ & 52.2\,/\,21.0 & \underline{0.59} & 2.72 
                                  & 62.0\,/\,30.2 & \textbf{0.58} & 0.48 
                                  & 29.6\,/\,11.0 & \underline{0.58} & 3.55 
                                  & 11.8\,/\,4.6 & \underline{0.56} & 188.4 
                                  & 13.2\,/\,5.4 & 0.60 & 61.9 \\
GeLLM$^4$O-C$_{\mathrm{Llama}}$ & 48.0\,/\,22.8 & 0.56 & 2.61 
                                & 53.8\,/\,24.8 & 0.56 & 0.47 
                                & 31.6\,/\,13.02 & 0.55 & 3.82 
                                & 10.2\,/\,4.6 & 0.54 & 40.2 
                                & 7.4\,/\,3.2 & 0.57 & \underline{259.2} \\

\rowcolor{gray!15} \multicolumn{16}{c}{\textit{RL Post-Training (\textsc{C-Moral}, Ours)}} \\
GRPO$_{\mathrm{Mistral}}$ & \underline{65.7}\,/\,36.4 & \textbf{0.60} & 3.22 
                          & \textbf{73.4}\,/\,\textbf{38.8} & 0.58 & \underline{0.51} 
                          & 42.4\,/\,20.8 & \textbf{0.59} & 3.54 
                          & \textbf{24.8}\,/\,\textbf{11.6} & \textbf{0.58} & 238.2 
                          & \textbf{38.2}\,/\,\textbf{18.0} & \underline{0.62} & 234.9 \\
GDPO$_{\mathrm{Mistral}}$ & \textbf{69.8}\,/\,\underline{40.2} & 0.58 &                                          \underline{4.05} 
                          & \underline{72.8}\,/\,\underline{37.0} & \underline{0.57} & \textbf{0.53} 
                          & \underline{44.4}\,/\,\underline{23.0} & 0.57 & \textbf{4.15} 
                          & \underline{22.6}\,/\,10.0 & \textbf{0.58} & \textbf{391.5} 
                          & \underline{25.6}\,/\,\underline{14.2} & \textbf{0.63} & 151.5 \\
GRPO$_{\mathrm{Llama}}$ & 57.8\,/\,26.6 & \underline{0.59} & 3.50
                        & 60.0\,/\,28.6 & \underline{0.57} & \underline{0.51}
                        & \textbf{45.2}\,/\,\textbf{25.0} & \underline{0.58} & 3.06
                        & 21.6\,/\,\underline{10.8} & \underline{0.56} & 256.2
                        & 16.6\,/\,9.2 & 0.60 & 249.5
\\
GDPO$_{\mathrm{Llama}}$ & 66.0\,/\,\textbf{44.8} & 0.57 & \textbf{4.43}
                        & 64.2\,/\,34.6 & \underline{0.57} & \underline{0.51}
                        & 43.7\,/\,22.7 & 0.57 & \underline{4.09}
                        & 18.5\,/\,9.9 & 0.56 & \underline{271.1}
                        & 24.4\,/\,15.1 & 0.60 & \textbf{271.6}
\\

\bottomrule
\end{tabular}
}
\caption{Detailed performance comparison on IND tasks. \textsc{C-Moral} consistently improves success rates over SFT baselines while preserving scaffold similarity. Bold denotes the best result in each column, and underlined denotes the second-best.}
\label{tab:ind_results}
\end{table*}

\begin{table*}[t]
\centering
\footnotesize
\setlength{\tabcolsep}{1.8pt}
\renewcommand{\arraystretch}{1.22}
\resizebox{\textwidth}{!}{
\begin{tabular}{@{}l ccc@{\hspace{4pt}} ccc@{\hspace{4pt}} ccc@{\hspace{4pt}} ccc@{\hspace{4pt}} ccc@{}}
\toprule
\multirow{2}{*}{\textbf{Model}} 
& \multicolumn{3}{c}{\textsc{CDE}} 
& \multicolumn{3}{c}{\textsc{ABMP}} 
& \multicolumn{3}{c}{\textsc{BCMQ}} 
& \multicolumn{3}{c}{\textsc{BDEQ}} 
& \multicolumn{3}{c}{\textsc{HLMPQ}} \\
\cmidrule(lr){2-4} \cmidrule(lr){5-7} \cmidrule(lr){8-10} \cmidrule(lr){11-13} \cmidrule(lr){14-16}
& S/SS$^{\uparrow}$ & Sim$^{\uparrow}$ & RI$^{\uparrow}$
& S/SS$^{\uparrow}$ & Sim$^{\uparrow}$ & RI$^{\uparrow}$
& S/SS$^{\uparrow}$ & Sim$^{\uparrow}$ & RI$^{\uparrow}$
& S/SS$^{\uparrow}$ & Sim$^{\uparrow}$ & RI$^{\uparrow}$
& S/SS$^{\uparrow}$ & Sim$^{\uparrow}$ & RI$^{\uparrow}$ \\
\midrule

\rowcolor{gray!15} \multicolumn{16}{c}{\textit{Supervised Fine-Tuning (SFT) Baseline}} \\
GeLLM$^4$O-C$_{\mathrm{Mistral}}$
& 4.8\,/\,1.4 & 0.57 & \textbf{120.1} 
& 45.4\,/\,22.8 & \textbf{0.60} & 3.03
& 40.8\,/\,18.6 & \underline{0.59} & 0.51
& 1.4\,/\,0.2 & \textbf{0.64} & 11.2
& 29.6\,/\,5.8 & \underline{0.58} & 1.49 \\

GeLLM$^4$O-C$_{\mathrm{Llama}}$
& 3.0\,/\,0.8 & \underline{0.60} & 18.0
& 52.6\,/\,23.2 & 0.56 & 1.79
& 38.8\,/\,20.0 & 0.56 & 0.52
& 1.4\,/\,0.6 & 0.59 & 10.9
& 30.2\,/\,6.4 & 0.54 & 1.31 \\

\rowcolor{gray!15} \multicolumn{16}{c}{\textit{RL Post-Training (\textsc{C-Moral}, Ours)}} \\
GRPO$_{\mathrm{Mistral}}$
& \textbf{7.6}\,/\,\textbf{2.6} & 0.58 & 78.7
& 71.8\,/\,47.6 & \underline{0.59} & \textbf{3.67}
& \underline{57.8}\,/\,\underline{34.2} & \textbf{0.60} & \textbf{0.55}
& \textbf{3.4}\,/\,1.2 & \underline{0.63} & 12.7
& \textbf{57.0}\,/\,\textbf{18.2} & \textbf{0.59} & \textbf{1.92} \\
GDPO$_{\mathrm{Mistral}}$
& \underline{6.4}\,/\,\underline{2.4} & 0.57 & \underline{114.1}
& 72.6\,/\,45.8 & 0.58 & \underline{3.40}
& \textbf{61.0}\,/\,\textbf{34.8} & 0.58 & \textbf{0.55}
& 2.8\,/\,1.2 & 0.62 & \underline{15.9}
& 48.8\,/\,14.8 & \textbf{0.59} & 1.90
\\
GRPO$_{\mathrm{Llama}}$
& 4.6\,/\,1.8 & \textbf{0.61} & 31.8
& \textbf{76.2}\,/\,\textbf{49.8} & \textbf{0.60} & 3.15
& 49.8\,/\,28.2 & 0.58 & 0.53
& \underline{3.0}\,/\,\textbf{1.6} & \underline{0.63} & 15.1
& \underline{52.6}\,/\,\underline{16.8} & \underline{0.58} & \textbf{1.92}
\\
GDPO$_{\mathrm{Llama}}$
& 4.5\,/\,2.0 & 0.59 & 72.4
& \underline{73.1}\,/\,\underline{47.7} & 0.58 & 3.24
& 53.3\,/\,33.9 & 0.58 & \underline{0.54}
& 2.9\,/\,\underline{1.5} & 0.62 & \textbf{17.4}
& 44.3\,/\,12.3 & \underline{0.58} & \underline{1.91}
\\
\bottomrule
\end{tabular}
}
\caption{Detailed performance comparison on OOD tasks. \textsc{C-Moral} consistently improves success rates over SFT baselines while preserving scaffold similarity. Bold denotes the best result in each column, and underlined denotes the second-best.}
\label{tab:ood_results}
\end{table*}

\section{Case Studies}
\subsection{BPQ Task}
BPQ (BBBP, PlogP, QED) involves diverse combinations of property-specific objectives across BBBP, PlogP, and QED, three key properties for CNS drug design. Each optimization task may require improving one or more properties while maintaining or further enhancing the others. Optimizing these diverse multi-objective combinations simulates the early-stage filtering and refinement of CNS-active hit compounds.

\begin{figure}[t]
    \centering

    \begin{subfigure}{0.9\linewidth}
        \centering
        \includegraphics[width=\linewidth]{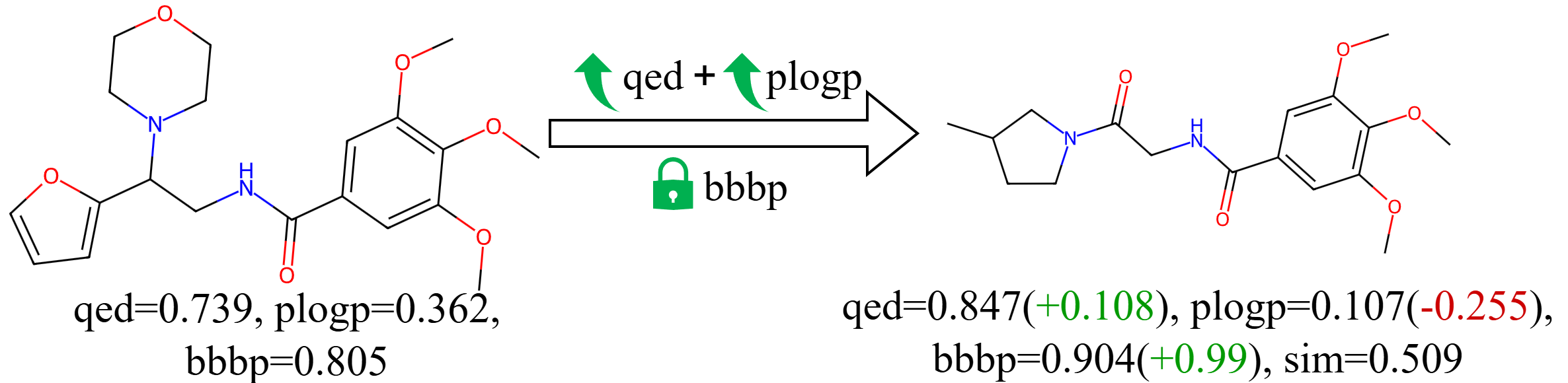}
        \caption{GeLLM$^4$O-C$_{\mathrm{Mistral}}$ optimization}
        \label{fig:bpq_mistral_gellmo}
    \end{subfigure}

    \vspace{0.8em}

    \begin{subfigure}{0.9\linewidth}
        \centering
        \includegraphics[width=\linewidth]{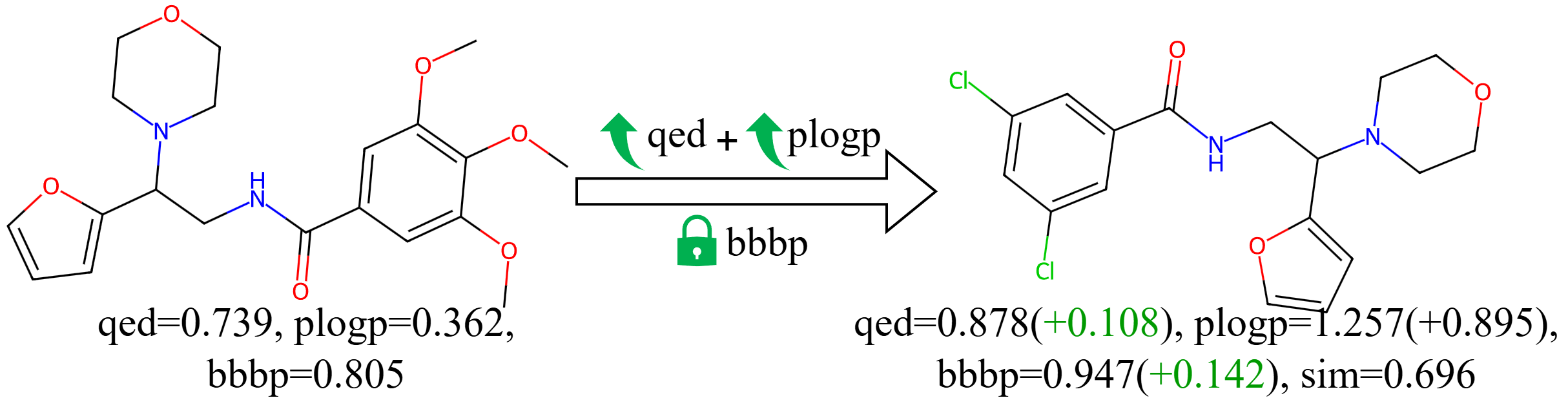}
        \caption{\textsc{C-Moral}-GRPO$_{\mathrm{Mistral}}$ optimization}
        \label{fig:bpq_mistral_grpo}
    \end{subfigure}

    \vspace{0.8em}

    \begin{subfigure}{0.9\linewidth}
        \centering
        \includegraphics[width=\linewidth]{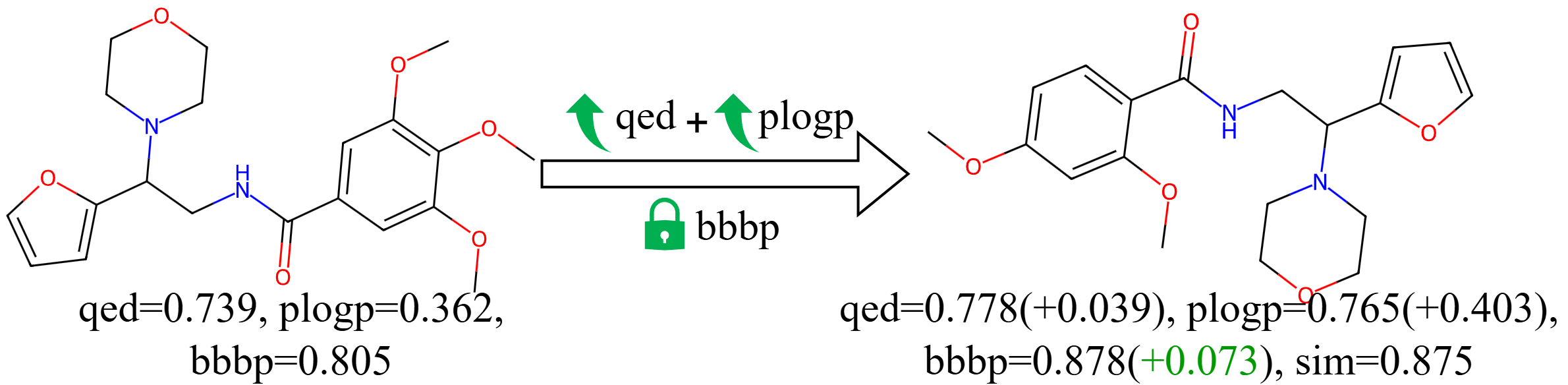}
        \caption{\textsc{C-Moral}-GDPO$_{\mathrm{Mistral}}$ optimization}
        \label{fig:bpq_mistral_gdpo}
    \end{subfigure}

    \caption{Optimization of different \textsc{Mistral}-based models on the \textsc{Bpq} task.}
    \label{fig:bpq_mistral_optimization}
\end{figure}

Figure~\ref{fig:bpq_mistral_optimization} presents one representative example from the BPQ task. Compared with the SFT baseline in Figure~\ref{fig:bpq_mistral_gellmo}, both RL post-trained models produce more favorable BPQ edits while better preserving meaningful structural motifs. GeLLM$^4$O-C$_{\mathrm{Mistral}}$ achieves the target improvement mainly through a relatively aggressive rewrite of the left-half structure, replacing the original peripheral heterocyclic region with a more compact motif while keeping only part of the right aromatic scaffold. In contrast, \textsc{C-Moral}-GRPO$_{\mathrm{Mistral}}$ in Figure~\ref{fig:bpq_mistral_grpo} performs a larger global restructuring, substantially changing both the aromatic core and the surrounding substituents; although this yields strong property improvement, it is less conservative in scaffold preservation. \textsc{C-Moral}-GDPO$_{\mathrm{Mistral}}$ in Figure~\ref{fig:bpq_mistral_gdpo}, however, makes more targeted local edits: it largely retains the original amide-linked aromatic core and the morpholine-containing motif, while modifying the peripheral substituents to improve QED and PlogP under the BBBP constraint. This qualitative example suggests that GDPO tends to achieve a better balance between property optimization and structural preservation, whereas GRPO explores more radical scaffold-level changes.

\section{Group-Based Policy Optimization Implementation}
\subsection{Group Relative Policy Optimization}
\label{appendix:grpo_implementation}
We provide the implementation details of GRPO used in our experiments.
Consider a mini-batch of prompts $\{x_i\}_{i=1}^{B}$.
For each prompt $x_i$, we sample a group of $G$ responses
\[
\mathcal{Y}_i = \{y_{i,1}, \dots, y_{i,G}\},
\qquad
y_{i,j} \sim \pi_{\Theta_{\mathrm{old}}}(\cdot \mid x_i).
\]
Assume there are $M$ reward functions.
The $m$-th reward of response $y_{i,j}$ is denoted by
\[
r_{i,j}^{(m)} = r_m(x_i, y_{i,j}),
\qquad
m = 1, \dots, M.
\]

In GRPO, the multiple reward dimensions are first linearly aggregated into a single scalar reward:
\[
r_{i,j}^{\mathrm{GRPO}} = \sum_{m=1}^{M} w_m\, r_{i,j}^{(m)},
\]
where $w_m \ge 0$ is the weight of the $m$-th reward.
In our default setting, we use $w_m = 1$ for all $m$ unless otherwise specified.

For each prompt $x_i$, we then compute the group mean and group standard deviation over the $G$ sampled responses:
\[
\mu_i
=
\frac{1}{G} \sum_{j=1}^{G} r_{i,j}^{\mathrm{GRPO}},
\]
\[
\sigma_i
=
\sqrt{
\frac{1}{G} \sum_{j=1}^{G}
\bigl(r_{i,j}^{\mathrm{GRPO}} - \mu_i\bigr)^2
}.
\]
The group-relative advantage is defined as
\[
A_{i,j}^{\mathrm{GRPO}}
=
\frac{r_{i,j}^{\mathrm{GRPO}} - \mu_i}{\sigma_i + \epsilon_{\mathrm{grp}}},
\]
where $\epsilon_{\mathrm{grp}}$ is a small constant for numerical stability.

For token $t$ in response $y_{i,j}$, the importance ratio is
\[
\rho_{i,j,t}(\Theta)
=
\frac{
\pi_{\Theta}(y_{i,j,t} \mid x_i, y_{i,j,<t})
}{
\pi_{\Theta_{\mathrm{old}}}(y_{i,j,t} \mid x_i, y_{i,j,<t})
}.
\]

The GRPO objective is
\[
\begin{split}
\mathcal{L}_{\mathrm{GRPO}}(\Theta)
=
\frac{1}{B} \sum_{i=1}^{B}
\frac{1}{G} \sum_{j=1}^{G}
\frac{1}{|y_{i,j}|} \\\sum_{t=1}^{|y_{i,j}|}
\min\Big(
\rho_{i,j,t}(\Theta)A_{i,j},
\\
\operatorname{clip}\big(\rho_{i,j,t}(\Theta),1-\epsilon_{\mathrm{clip}},1+\epsilon_{\mathrm{clip}}\big)
A_{i,j}
\Big).
\end{split}
\]

If a KL regularization term is used in practice, it can be added in the standard way.
Compared with GDPO, GRPO performs normalization only after collapsing all reward dimensions into a single scalar reward, rather than normalizing each reward dimension separately before aggregation.

\subsection{Group reward-Decoupled Normalization
Policy Optimization}
\label{appendix:gdpo_implementation}

We provide the implementation details of GDPO used in our experiments.
Consider a mini-batch of prompts $\{x_i\}_{i=1}^{B}$.
For each prompt $x_i$, we sample a group of $G$ responses
\[
\mathcal{Y}_i = \{y_{i,1}, \dots, y_{i,G}\}, \qquad
y_{i,j} \sim \pi_{\Theta_{\mathrm{old}}}(\cdot \mid x_i).
\]
Assume there are $M$ reward functions.
The $m$-th reward of response $y_{i,j}$ is denoted by
\[
r_{i,j}^{(m)} = r_m(x_i, y_{i,j}), \qquad m = 1, \dots, M.
\]

\paragraph{Step 1: Group-wise decoupled normalization.}
For each prompt $x_i$ and each reward dimension $m$, we first compute the
group mean and group standard deviation over the $G$ sampled responses:
\[
\mu_i^{(m)} = \frac{1}{G} \sum_{j=1}^{G} r_{i,j}^{(m)},
\]
\[
\sigma_i^{(m)} =
\sqrt{
\frac{1}{G} \sum_{j=1}^{G} \bigl(r_{i,j}^{(m)} - \mu_i^{(m)}\bigr)^2
}.
\]
Then the reward-specific normalized advantage is
\[
A_{i,j}^{(m)} =
\frac{r_{i,j}^{(m)} - \mu_i^{(m)}}{\sigma_i^{(m)} + \epsilon_{\mathrm{grp}}}.
\]

\paragraph{Step 2: Aggregate the decoupled advantages.}
We sum the normalized advantages from all reward dimensions:
\[
\tilde{A}_{i,j} = \sum_{m=1}^{M} w_m A_{i,j}^{(m)},
\]
where $w_m \ge 0$ is the weight of the $m$-th reward.
In our default setting, we use $w_m = 1$ for all $m$ unless otherwise specified.

\paragraph{Step 3: Batch-wise normalization (BN).}
To keep the numerical scale of advantages stable as the number of rewards increases,
we further normalize $\tilde{A}_{i,j}$ over all responses in the current mini-batch.
Let
\[
\mu_{\mathcal{B}} =
\frac{1}{BG} \sum_{i=1}^{B} \sum_{j=1}^{G} \tilde{A}_{i,j},
\]
\[
\sigma_{\mathcal{B}} =
\sqrt{
\frac{1}{BG} \sum_{i=1}^{B} \sum_{j=1}^{G}
\bigl(\tilde{A}_{i,j} - \mu_{\mathcal{B}}\bigr)^2
}.
\]
The final GDPO advantage is
\[
\hat{A}_{i,j} =
\frac{\tilde{A}_{i,j} - \mu_{\mathcal{B}}}{\sigma_{\mathcal{B}} + \epsilon_{\mathrm{bn}}}.
\]
This step is crucial in the original paper. After testing, we find it also important for stable training.

\paragraph{Step 4: Policy optimization objective.}
We then use the final normalized advantage $\hat{A}_{i,j}$ in a clipped policy
optimization objective.
For token $t$ in response $y_{i,j}$, define the importance ratio as
\[
\rho_{i,j,t}(\Theta)
=
\frac{
\pi_{\Theta}(y_{i,j,t} \mid x_i, y_{i,j,<t})
}{
\pi_{\Theta_{\mathrm{old}}}(y_{i,j,t} \mid x_i, y_{i,j,<t})
}.
\]
The GDPO objective is
\[
\begin{aligned}
\mathcal{L}_{\mathrm{GDPO}}(\Theta)
&=
\frac{1}{B} \sum_{i=1}^{B}
\frac{1}{G} \sum_{j=1}^{G}
\frac{1}{|y_{i,j}|} \sum_{t=1}^{|y_{i,j}|}
\min\!\Bigl(
\rho_{i,j,t}(\Theta)\,\hat{A}_{i,j}, \\
&\qquad\qquad
\mathrm{clip}\bigl(
\rho_{i,j,t}(\Theta),\, 1-\epsilon_{\mathrm{clip}},\, 1+\epsilon_{\mathrm{clip}}
\bigr)\hat{A}_{i,j}
\Bigr).
\end{aligned}
\]

If a KL regularization term is used in practice, it can be added in the standard way.
The key difference from multi-reward GRPO is that GDPO normalizes each reward
\emph{before} aggregation, and then applies an additional batch-wise normalization
to the aggregated advantage.

\end{document}